\documentclass{article}

\PassOptionsToPackage{numbers, compress}{natbib}

\usepackage[preprint]{neurips_2025}

\usepackage[utf8]{inputenc} 
\usepackage[T1]{fontenc}    
\usepackage{hyperref}       
\usepackage{url}            
\usepackage{booktabs}       
\usepackage{amsfonts}       
\usepackage{enumitem}       
\usepackage{nicefrac}       
\usepackage{microtype}      
\usepackage{xcolor}         
\usepackage{amsmath}

\usepackage{multirow}
\usepackage{array}
\usepackage{arydshln}
\usepackage{graphicx}
\usepackage{subcaption}
\usepackage{colortbl}
\usepackage{array}
\usepackage{wrapfig}

\title{TACO: Think-Answer Consistency for Optimized Long-Chain Reasoning and Efficient Data Learning via Reinforcement Learning in LVLMs}

\author{%
  Zhehan Kan\thanks{Equal contribution. Zhehan Kan works done during his internship at Tencent YouTu Lab.}\hspace{0.45em}\textsuperscript{1}\textsuperscript{2}%
  \quad Yanlin Liu\footnotemark[1]\hspace{0.45em}\textsuperscript{1}%
  \quad Kun Yin\footnotemark[1]\hspace{0.45em}\textsuperscript{2}%
  \quad Xinghua Jiang\textsuperscript{2}%
  \quad Xin Li\textsuperscript{2}%
  \quad Haoyu Cao\textsuperscript{2}\\[0.5em]
  \textbf{Yinsong Liu\textsuperscript{2}%
    \quad Deqiang Jiang\textsuperscript{2}%
    \quad Xing Sun\textsuperscript{2}%
    \quad Qingmin Liao\textsuperscript{1}%
    \quad Wenming Yang\textsuperscript{1}}}

\begin{document}

\maketitle

\vspace{-3em}
\begin{center}
  \textsuperscript{1}Tsinghua University \quad
  \textsuperscript{2}Tencent YouTu Lab
\end{center}

\begin{abstract}
  DeepSeek R1 has significantly advanced complex reasoning for large language models (LLMs). While recent methods have attempted to replicate R1’s reasoning capabilities in multimodal settings, they face limitations, including inconsistencies between reasoning and final answers, model instability and crashes during long-chain exploration, and low data learning efficiency. To address these challenges, we propose TACO, a novel reinforcement learning algorithm for visual reasoning. Building on Generalized Reinforcement Policy Optimization (GRPO), TACO introduces Think-Answer Consistency, which tightly couples reasoning with answer consistency to ensure answers are grounded in thoughtful reasoning. We also introduce the Rollback Resample Strategy, which adaptively removes problematic samples and reintroduces them to the sampler, enabling stable long-chain exploration and future learning opportunities. Additionally, TACO employs an adaptive learning schedule that focuses on moderate difficulty samples to optimize data efficiency. Furthermore, we propose the Test-Time-Resolution-Scaling scheme to address performance degradation due to varying resolutions during reasoning while balancing computational overhead. Extensive experiments on in-distribution and out-of-distribution benchmarks for REC and VQA tasks show that fine-tuning LVLMs leads to significant performance improvements.
\end{abstract}

\section{Introduction}

\begin{figure*}[t]
    \centering
    \includegraphics[width=390pt,height=250pt]{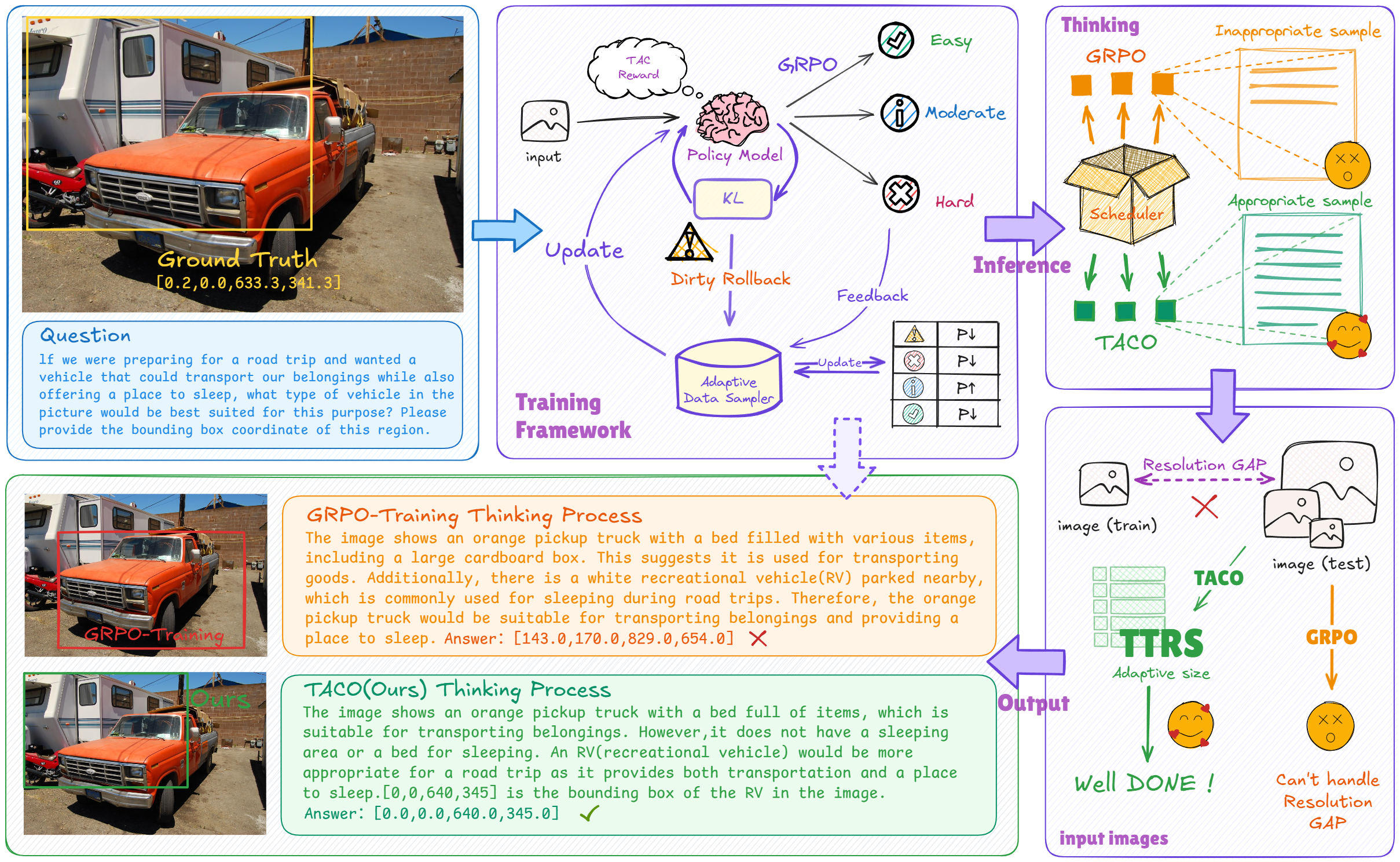}
    \caption{An example of a REC task: Illustrating (center) an enhanced GRPO-based learning loop with TAC, RRS, and ADS. Qualitative comparison between GRPO and TACO (top right) demonstrates TACO's superior inference sampling. TTRS module (bottom right) effectively addresses resolution gaps between training and testing images. TACO's accurate output(on REC) and reasoning process closely mirroring the ground truth are exemplified by a visual reasoning task (left).} 
    \label{fig:intro}
    \vspace{-5mm}
\end{figure*}

Large Vision-Language Models (LVLMs) have emerged as powerful tools for the intelligent processing and understanding of multimodal data, such as images and text \cite{zhang2024visionlanguagemodelsvisiontasks}. These models have evolved significantly, progressing from early stages of simple feature fusion \cite{xu2016showattendtellneural, agrawal2016vqavisualquestionanswering} to the current state, which involves complex end-to-end training and multi-stage instruction fine-tuning \cite{achiam2023gpt, bai2025qwen2, lu2024deepseekvlrealworldvisionlanguageunderstanding, chen2024internvl}. The predominant training strategy today is a combination of ``Pre-training + Multi-stage Alignment/Post-training'' \cite{sun20233d}. Despite these advancements, LVLMs still face challenges in comprehending instructions and performing visual reasoning tasks effectively. To address these deficiencies, Supervised Fine-Tuning (SFT) has been used to improve model capabilities \cite{liu2023visual, dai2023instructblipgeneralpurposevisionlanguagemodels, zhu2023minigpt4enhancingvisionlanguageunderstanding}. However, SFT often results in ``catastrophic forgetting'', a phenomenon where the model loses previously learned information when updated, leading to poor generalization in fine-tuned models. To overcome these issues, Reinforcement Learning from Human Feedback (RLHF) became a useful technique for aligning LVLMs with human expectations \cite{xiao2025detecting}. Nevertheless, the substantial human labor costs and potential annotator biases associated with RLHF present challenges for its scalability \cite{xu2025rlthftargetedhumanfeedback}. 

DeepSeek-R1 model \cite{guo2025deepseek} provided a new direction for enhancing LVLMs' complex reasoning capabilities. It shifts the focus from a reliance on ``imitation learning'' to one that emphasizes ``problem solving'', addressing many of the inefficiencies, instabilities, and high annotation costs associated with traditional RL approaches. Following R1's success with Group Relative Policy Optimization (GRPO) in improving reasoning for large language models (LLMs), researchers have begun applying it to the more challenging field of LVLMs. While GRPO has shown promise in areas like mathematics and programming \cite{zhang2025grpoleaddifficultyawarereinforcementlearning, wei2025swe}, visual reasoning, particularly tasks grounded in perception, is becoming a key area for evaluating LVLMs' reasoning abilities \cite{xiao2024visualgroundingsurvey}. Early efforts, such as VLM-R1 \cite{shen2025vlm}, have made progress, but significant challenges remain.

The substantial human labor costs and potential biases in RLHF became bottlenecks \cite{xu2025rlthftargetedhumanfeedback}. The DeepSeek-R1 model \cite{guo2025deepseek} offers a new approach by shifting from ``imitation'' learning to ``problem-solving'', addressing efficiency and annotation cost challenges of traditional RL. Building on R1's success with Group Relative Policy Optimization (GRPO) in LLM reasoning tasks, researchers are now exploring its use in LVLMs. While progress has been made in tasks like mathematics and code \cite{zhang2025grpoleaddifficultyawarereinforcementlearning, wei2025swe}, perception-grounding tasks \cite{xiao2024visualgroundingsurvey} are more crucial for evaluating LVLM reasoning. Efforts like VLM-R1 \cite{shen2025vlm} face the following issues: 1) \textbf{Invalid Reasoning:} GRPO training generates a thinking process but fails to map it to the answer, causing think-answer inconsistency and short CoT. 2) \textbf{Long Chain Exploration Collapse:} As response length grows, the model becomes fragile and collapses early in long-chain exploration. 3) \textbf{Inefficient Learning:} RL sample selection is crucial, but random sampling and offline learning hinder appropriate knowledge acquisition, while online learning introduces bias, leading to local optima. 4) \textbf{Training-Testing Resolution Gap:} RL training requires high GPU memory, and using compressed images for training creates a resolution gap between training and testing, negatively impacting performance.

To address prior issues (Fig.~\ref{fig:intro}), we propose TACO, built on GRPO. It uses \textbf{Think-Answer Consistency (TAC)} for coherent responses. When TAC reasons, \textbf{Rollback Resample Strategy (RRS)} prevents model collapse from gradients of temporary ``dirty samples'', aiding future learning. \textbf{Adaptive Difficulty Sampling (ADS)} boosts efficiency by focusing on moderate-difficulty samples. \textbf{Test-Time-Resolution-Scaling (TTRS)} bridges train-test resolution gaps via multi-scale test sampling. TACO applies to various LVLMs. Experiments with Qwen 2.5-VL-3B on 15 in/out-of-domain REC/VQA test sets show TACO improves performance, generalization, and versatility.

The contributions of this paper are as follows: 1) We propose TACO, a novel RL algorithm for visual reasoning in LVLMs that ensures Think-Answer consistency, stabilizes early long-chain exploration, and enhances learning efficiency with adaptive resampling. 2) We identify the resolution distribution gap as a key challenge in visual reasoning for RL and introduce a multi-scale sampling method during testing to improve performance without additional training or reasoning overhead. 3) Extensive experiments show that TACO significantly enhances performance in both in-domain and out-of-domain REC and VQA tasks, with strong generalization and versatility.

\section{Related Work}
\label{gen_inst}

\textbf{Large Vision-Language Models (LVLMs)}. LVLMs bridge vision and language. Core advances include large-scale contrastive pre-training for joint embeddings (e.g., CLIP \cite{radford2021learning}) and LLM-style instruction tuning for enhanced visual dialogue/reasoning (e.g., LLaVA \cite{liu2023visual}). Dealing with varied image sizes is key. Dynamic methods (AnyRes \cite{chen2024internvl}; QwenVL techniques \cite{bai2023versatile}) aid input flexibility. However, complex reasoning and generalization remain tough.

\textbf{Reinforcement Learning (RL) in LVLMs}. RL offers a compelling way to enhance reasoning, building on language successes like RL's efficacy on logical tasks~\cite{OpenAI_o1_Preview_2024} and GRPO enabling direct reasoning optimization (potentially bypassing SFT, DeepSeek-R1~\cite{guo2025deepseek}). For multimodal RL, however, addressing cross-modal consistency and stability is key. Efforts in this area include developing specialized reasoning datasets with formalized visual inputs (R1-OneVision \cite{yang2025r1}), successfully porting RL algorithms like GRPO to VLM training (R1-V, Visual-RFT, VLM-R1 \cite{R1-V, liu2025visual, shen2025vlm}), and introducing mechanisms like verifiable rewards \cite{liu2025visual}. Intriguingly, applying RL directly to base VLMs has been shown to induce significant performance jumps or ``visual epiphanies'' (VisualThinker-R1-Zero \cite{zhou2025r1}), with related work observing correlations between response characteristics like length and reasoning improvements under RL optimization (MMEureka \cite{meng2025mmeurekaexploringfrontiersmultimodal}).

VLM-R1 \cite{shen2025vlm} applies GRPO to visual reasoning tasks, showcasing RL's advantages over SFT in generalization. However, it encounters challenges such as reasoning inconsistencies, model instability during long-chain exploration, low data learning efficiency, and the training-testing resolution gap. TACO addresses these by coupling reasoning with answer consistency, dynamically rolling back temporary ``dirty samples'', focusing on moderate samples, and employing multi-scale resolution ensembles during testing. These enhancements enable LVLMs to achieve high-quality learning, boosting their reasoning and generalization capabilities in visual tasks through RL.

\section{Method}
\label{headings}

\begin{figure*}[t]
    \centering
    
    \includegraphics[width=340pt, height=180pt]{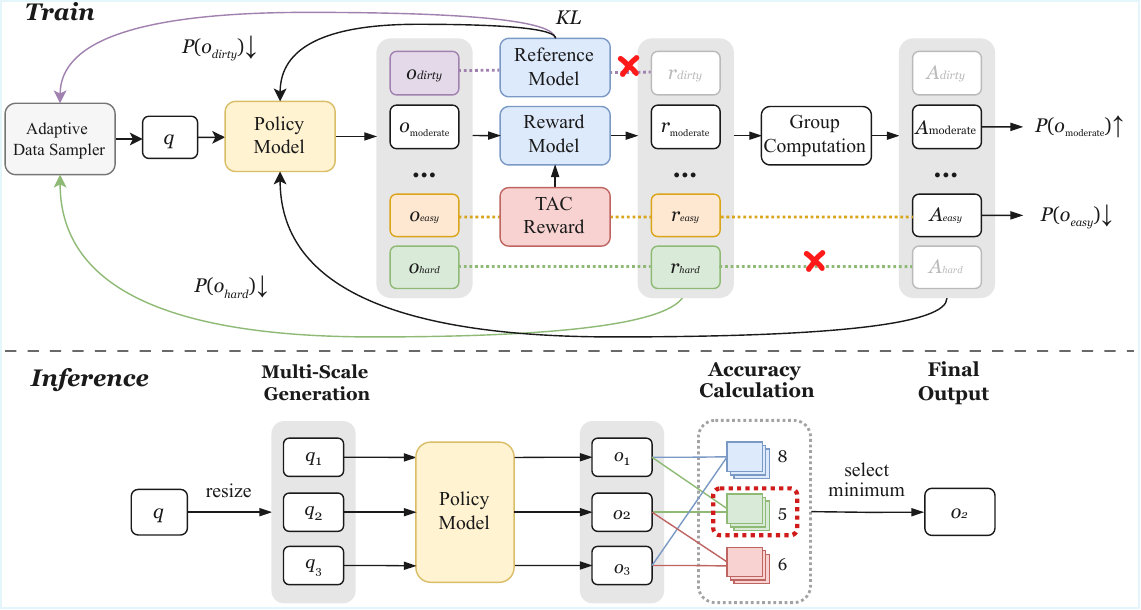}
    \caption{During training, the TAC reward ensures consistent Think-Answer output. Samples are initially given equal sampling rates, with temporary ``dirty samples'' identified by the KL divergence between the current policy $\pi_\theta$ and the reference policy $\pi_{\text{ref}}$. Their gradients are masked, and sampling rates are reduced to stabilize long-chain exploration and allow future resampling. Samples are classified as easy, moderate, or hard based on accuracy rewards, with easy samples rarely resampled, hard samples slightly reduced, and moderate samples increased for focused learning. Multiple scale resolutions are sampled during reasoning, and the answer with the least intersection is selected. Since the number of samples is small and the test image is compressed, reasoning time remains nearly. 
    }
    \label{fig:overview}
    \vskip -0.1in 
    \vspace{-3mm}
\end{figure*}

As shown in Figure~\ref{fig:overview}, TACO, built upon GRPO, incorporates four novel components to enhance reasoning ability and learning efficiency: 1) Think-Answer Consistency (TAC), which ensures coherent alignment between the model's reasoning process, its final answer, and the ground truth; 2)  Rollback Resample Strategy (RRS), which improves training stability by managing temporary ``dirty samples'' that cause mutation shifts between the current step and the base model; 3) Adaptive Difficulty Sampling (ADS) with offline curation to optimize learning efficiency; and 4) Test-Time Resolution Scaling (TTRS) to bridge the gap between training and testing data. These components work synergistically to foster robust, accurate, and scalable visual reasoning.

\subsection{Preliminary}

\paragraph{Group Relative Policy Optimization}

Group Relative Policy Optimization (GRPO) enhances PPO \cite{schulman2017proximal} by eliminating the critic component. For input $q$, GRPO samples $N$ responses $\{o_1, o_2, \ldots, o_N\}$ from policy $\pi_{\theta}$, computing rewards $r_i = R(q, o_i)$. Relative performance is evaluated using advantage values $A_i$
, which standardizes rewards without requiring a separate value function.
The policy $\pi_{\theta}$ is updated by optimizing the GRPO objective:
\begin{equation}
    J_{\text{GRPO}}(\theta) = \mathbb{E}_{\{o_i\}_{i=1}^N \sim \pi_{\theta_{\text{old}}}(q)} \left[ \frac{1}{N} \sum_{i=1}^N \{\min(s_1 \cdot A_i, s_2 \cdot A_i) \} - \beta D_{\text{KL}}[\pi_{\theta} \| \pi_{\text{ref}}] \right],
\end{equation}
where $s_1 = \frac{\pi_{\theta}(o_i|q)}{\pi_{\theta_{\text{old}}}(o_i|q)}$ and $s_2 = \text{clip}\left(\frac{\pi_{\theta}(o_i|q)}{\pi_{\theta_{\text{old}}}(o_i|q)}, 1-\epsilon, 1+\epsilon\right)$.
The term $D_{\text{KL}}[\pi_{\theta} \| \pi_{\text{ref}}]$ represents the Kullback-Leibler divergence between the current and reference policies, weighted by $\beta$.

\paragraph{Referring Expression Comprehension and Visual Question Answering.} Referring Expression Comprehension (REC) is a multimodal task enabling machines to localize target objects or regions within a visual scene based on a natural language expression \cite{qiao2020referring}. Unlike traditional object detection, REC requires complex instructions and enhanced visual perception, making it valuable for applications like human-centric scenarios, autonomous driving, and medical image analysis \cite{he2023grec}. Visual Question Answering (VQA) tasks, in contrast, involve generating accurate natural language answers based on an image and a question \cite{antol2015vqa, srivastava2021visual}. Successful completion of VQA demands capabilities in object recognition, attribute understanding, and relational analysis. We extend experiments on these tasks to validate the robust reasoning and generalization capabilities of TACO.

\subsection{Think-Answer Consistency (TAC)}
\label{sec: tac}

To address the issue of inconsistencies between the reasoning process and the final answer in LVLMs during visual reasoning tasks, and to ensure that the model generates answers with careful reasoning, we propose the Think-Answer Consistency (TAC) reward. The core idea of TAC is to directly supervise the alignment between the model's reasoning process ($\mathit{Think}$) and the final answer ($\mathit{Answer}$) with the Ground Truth ($\mathit{GT}$), thus preventing the model from ``bypassing'' the reasoning process and producing lazy outputs. The general form of the TAC reward can be expressed as:
\begin{equation}
    R_{TAC} = \mathit{f}(\mathit{Think}, \mathit{Answer}, \mathit{GT}),
\end{equation}
where $\mathit{f}$ is a metric function designed according to the specific task, used to evaluate the degree of alignment between $\mathit{Think}$, $\mathit{Answer}$, and $\mathit{GT}$. 

In the REC task, the model's objective is to locate the target object in an image based on a given referring expression, typically represented as a Bounding Box (BBox). To decouple the reasoning and the final output, we prompt the model with: \textit{``First output the thinking process, then summarize the answer in <think> </think> tags, and output the final answer in <answer> </answer> tags.''} We then extract the model’s thinking process ($\mathit{Think}_{BBox}$) and its final answer ($\mathit{Answer}_{BBox}$). Using the thinking process, the final answer, and the ground truth target ($\mathit{GT}_{BBox}$), we calculate the Intersection over Union (IoU) of these three BBoxes. This IoU serves as the TAC reward in the REC task. The reward is $R^{REC} = R_{acc}^{REC} + R_{format}$, with $R_{acc}^{REC}$ given by:
  \begin{equation}  
  \begin{aligned}  
  R_{acc}^{REC} = R_{TAC}^{REC} &= IoU(\mathit{Think}_{BBox}, \mathit{Answer}_{BBox}, \mathit{GT}_{BBox}) \\  
  &=  \frac{Area(BBox_1 \cap BBox_2 \cap BBox_3)}{Area(BBox_1 \cup BBox_2 \cup BBox_3)},  
  \end{aligned}  
  \end{equation}  
  where $IoU(BBox_1, BBox_2, BBox_3)$ is calculated as the ratio of the intersection area of these three BBoxes to their union area, and the $R_{format}$ is defined as $(<think>...</think><answer>...</answer>)$. We use $R_{TAC}^{REC}$ as both an accuracy reward and a consistency reward in the REC task. In this way, we rigorously supervise the alignment between the thinking process, the answer, and the ground truth, effectively reinforcing the dependency between the thinking process and the answer.

In VQA tasks, models generate text-based answers from image-question pairs. However, discrete rule-based rewards in VQA (such as selection or judgment) introduce randomness, hindering explicit consistency supervision. To address this, we employ an external supervisor model (S) with strong reasoning capabilities to assess both the consistency and correctness of the generated thinking process (T) and answer (A). Given a question Q, the supervisor S evaluates the model-generated T and A. In this work, we use Qwen 2.5-VL-32B as S, prompting the model with: \textit{``As a text comprehension expert, evaluate the semantic consistency between automatically extracted answers based on the given corpus, questions, and reference answers, outputting a similarity score within the [0,1] range. Output ONLY the score.''} The reward in VQA, denoted as $R^{VQA} = R_{TAC}^{VQA} + R_{acc}^{VQA} + R_{format}$, is based on the following:
\begin{equation}
 \begin{aligned}
  R_{TAC}^{VQA} = S(Q, T, GT),
 \end{aligned}
\end{equation}
where $R_{acc}^{VQA}$ represents the accuracy reward: in closed-ended scenarios, it compares the model’s answer with the reference, returning 1 or 0 based on correctness, and in open-ended scenarios, it calculates the Edit distance between the model’s answer and the ground truth.

\subsection{Rollback Resample Strategy (RRS)}
\label{sec:RRS}

\begin{figure*}[t]
    \centering
    
    \begin{subfigure}[b]{0.48\linewidth} 
        \includegraphics[width=\linewidth]{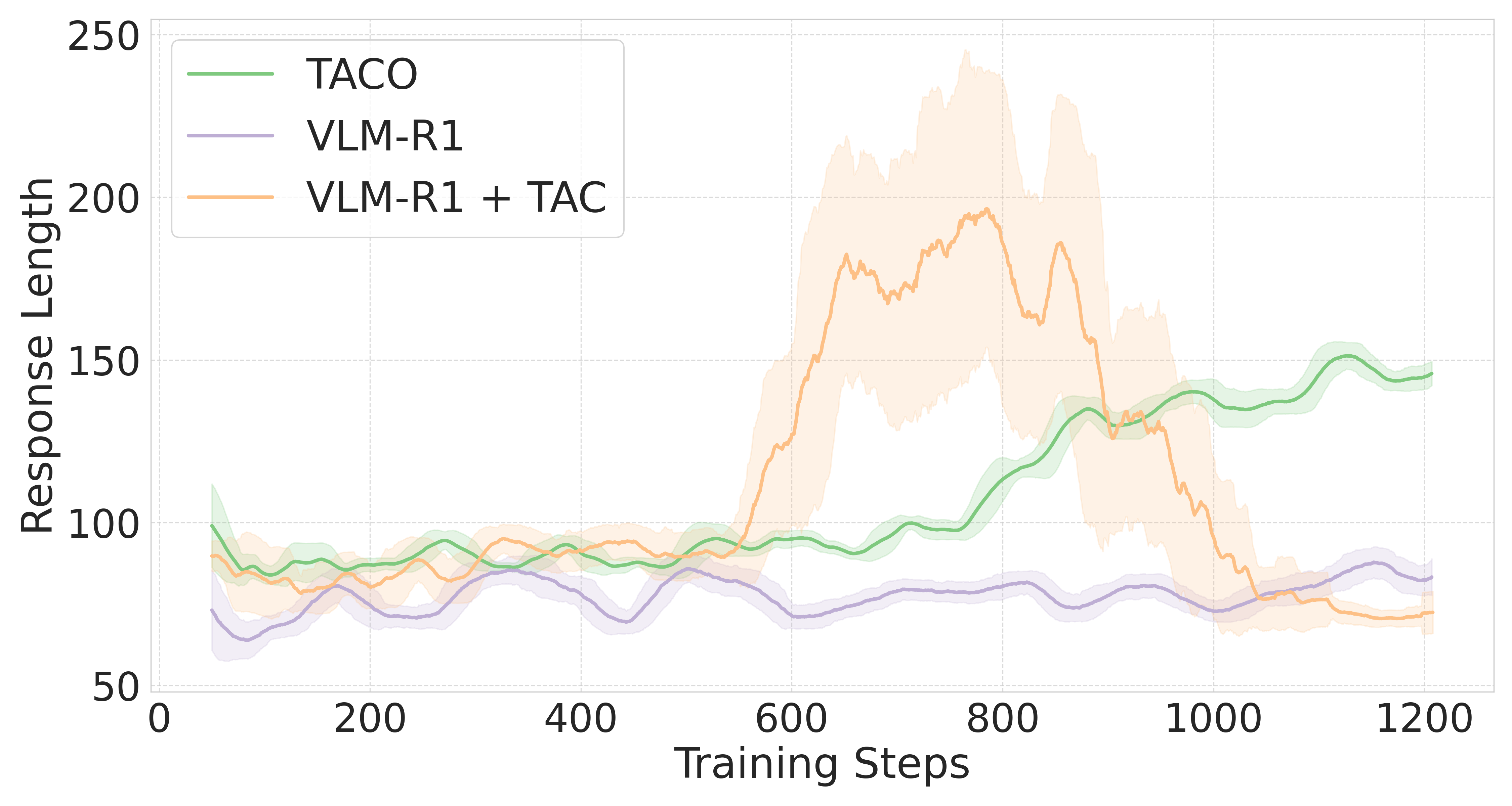}
        \caption{Response length comparison.} 
        \label{fig:sub_com_len_comparison_dr}
    \end{subfigure}
    \hfill 
    \begin{subfigure}[b]{0.48\linewidth} 
        \includegraphics[width=\linewidth]{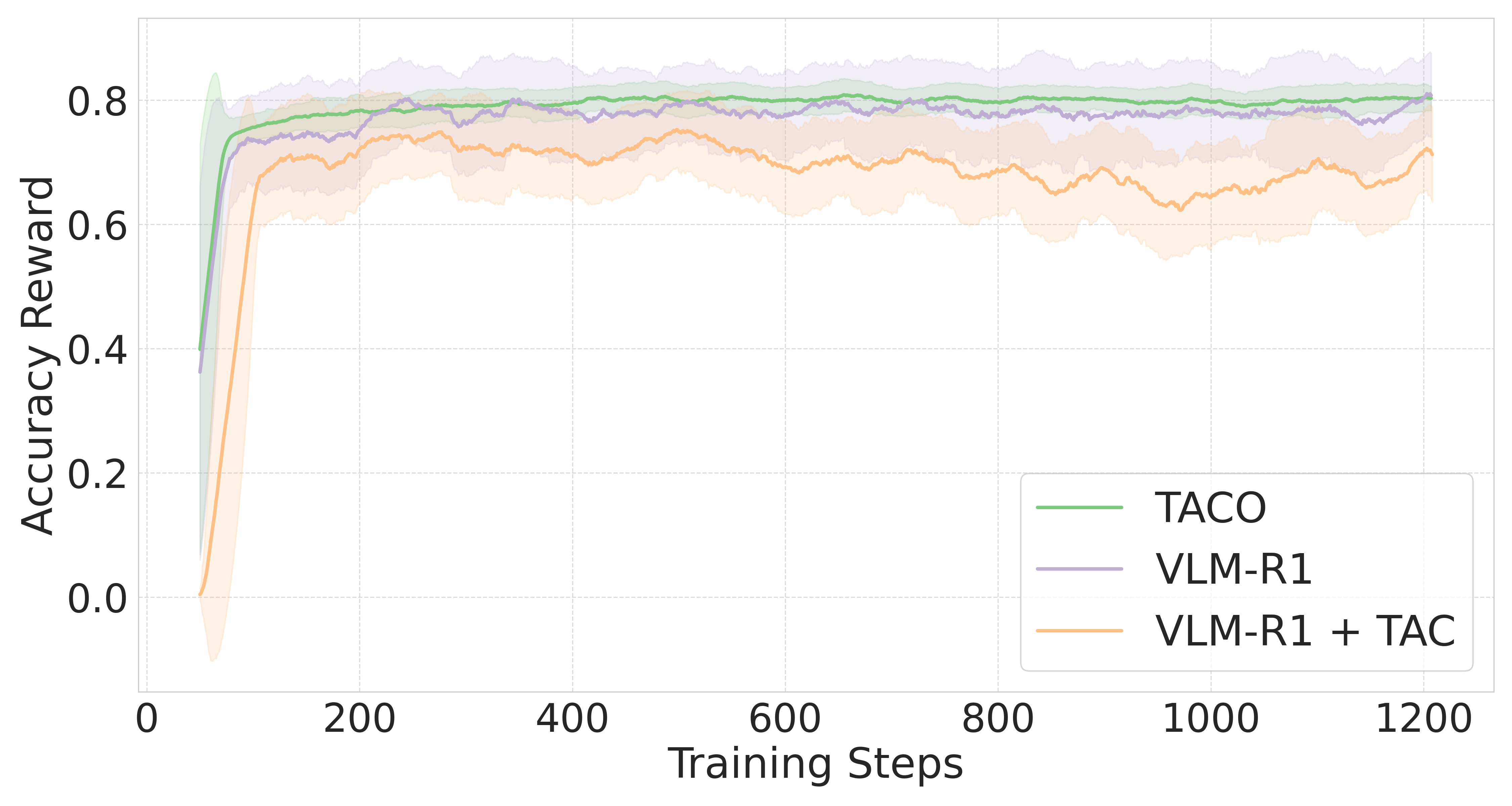}
        \caption{Training accuracy (IoU Reward) comparison.} %
        \label{fig:sub_test_result_comparison_dr}
    \end{subfigure}
    \begin{subfigure}[b]{0.48\linewidth} 
        \includegraphics[width=\linewidth]{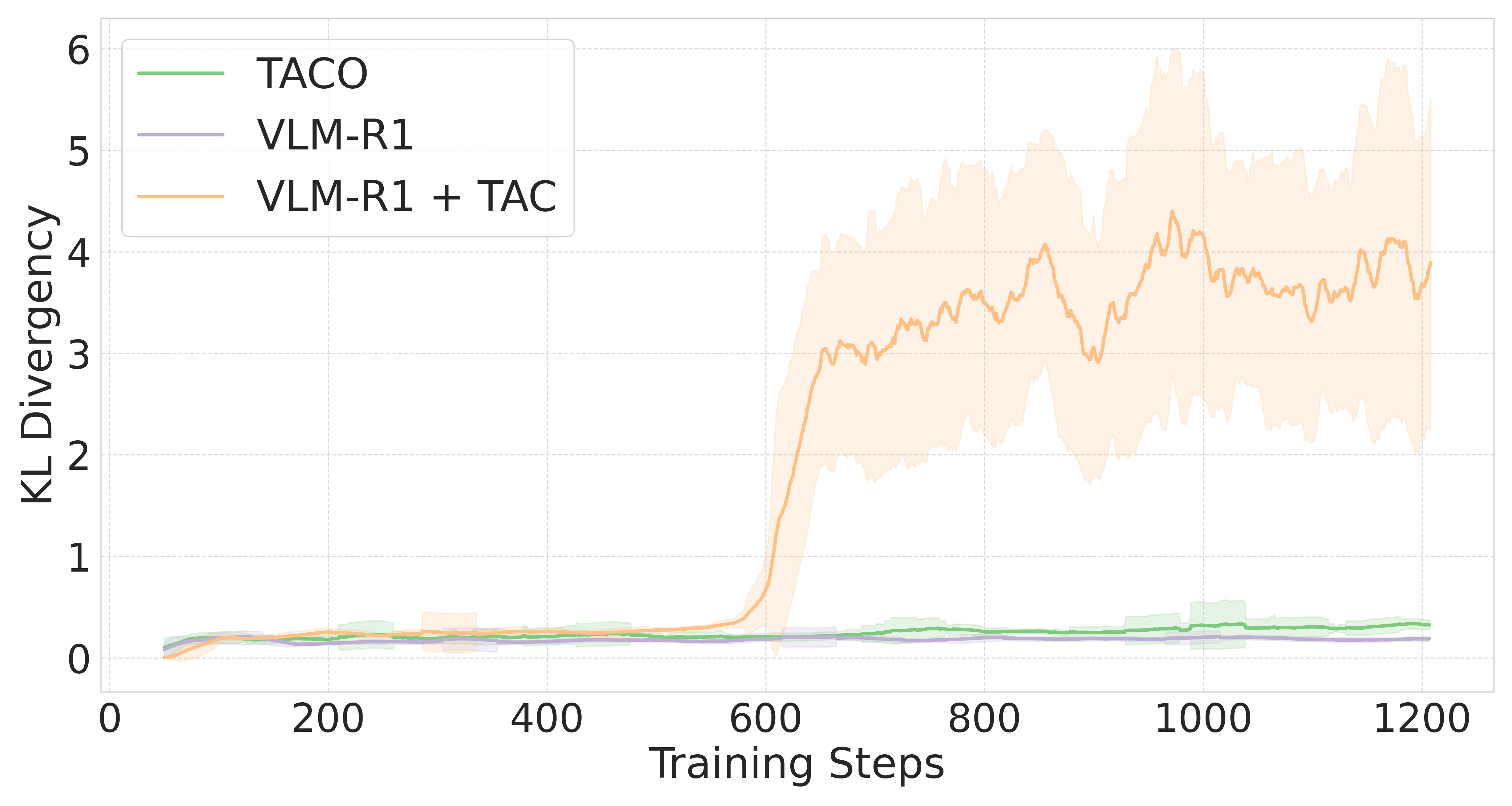}
        \caption{KL divergence comparison.} 
        \label{fig:sub_com_len_comparison_KL}
    \end{subfigure}
    \hfill 
    \begin{subfigure}[b]{0.48\linewidth} 
        \includegraphics[width=\linewidth]{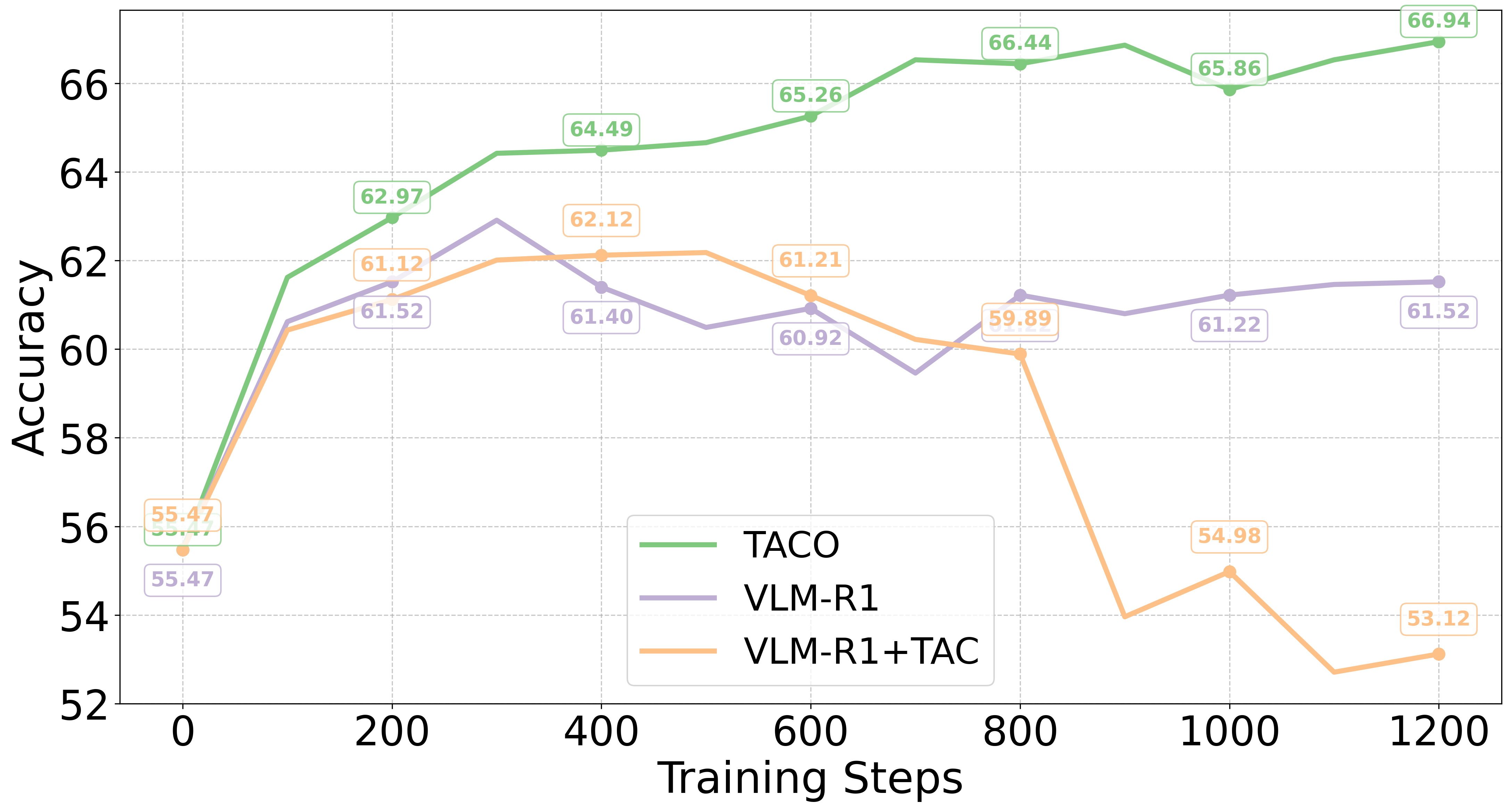}
        \centering
        \caption{LISA test accuracy comparison.}
        \label{fig:LISA_test accuracy comparison}
    \end{subfigure}
    
    \caption{Effectiveness of the Think-Answer Consistency (TAC) reward, comparing TACO, VLM-R1, and VLM-R1 + TAC. The subplots illustrate TAC's influence on: (a) response length evolution; (b) training accuracy (IoU reward) and the critical reasoning-answer alignment; (c) policy stability, tracked via KL divergence; and (d) Performance on the LISA test set.}

    \label{fig:dirty_rollback_comparison_overall} 
    
    \vskip -0.2in 
\end{figure*}

Long Chains of Thought (CoT) are essential for complex reasoning, but traditional rule-based accuracy rewards (e.g., the purple line in Figure~\ref{fig:dirty_rollback_comparison_overall}) neglecting think-answer consistency lead to short responses and ineffective reasoning (see Section~\ref{sec: tac}). TAC, shown in the orange line of Figure~\ref{fig:dirty_rollback_comparison_overall}, activates reasoning and increases CoT length by ensuring consistency between thoughts and answers. However, after initial growth, response length drops sharply. Metrics in Figure~\ref{fig:dirty_rollback_comparison_overall} collapse synchronously, indicating that this occurs during long-chain RL exploration when the model is fragile. Complex samples can temporarily become ``dirty samples'', creating gaps between current ($\pi_\theta$) and reference ($\pi_{\text{ref}}$) policies. This spikes KL divergence $D_{\text{KL}}(\pi_\theta \| \pi_{\text{ref}})$, which then dominates the loss, voiding reward-based learning. The model then outputs repetitive answers and garbled code (examples in supplement), slashing response length and accuracy reward, ultimately leading to collapse.

To enable effective long CoT and stimulate the model's reasoning ability across visual tasks, we designed the Rollback Resample Strategy (RRS), which ensures stability during long-chain reasoning by dynamically managing temporary ``dirty samples''. We define the divergence between the current policy $\pi_\theta$ and the reference policy $\pi_{\text{ref}}$ as  $D_{KL}(x) = D_{KL}(\pi_\theta(\cdot|x) \| \pi_{ref}(\cdot|x))$. Initially, each input sample $i$ is assigned a sampling rate $P(i)$ of 1.0. Then, by calculating $D_{KL}(i)$ samples into normal $i_n$ and dirty $i_d$, as defined below:
\begin{equation}
i =
\begin{cases}
    i_d & \text{if } D_{KL}(\pi_\theta(\cdot|x) \| \pi_{ref}(\cdot|x)) > \kappa \\
    i_n & \text{otherwise},
\end{cases}
\label{eq:sample_devide}
\end{equation}
where $\kappa$ represents a hyperparameter, set to 0.5. For normal sample $i_n$, we backpropagate the gradients and maintain the sampling rate. For the ``dirty sample'' $i_d$, RRS applies two measures: \textbf{Gradient Masking:} The gradient of the ``dirty sample'' $i_d$ generated by the policy $\pi_\theta$ is masked from gradient computations during the current training step. \textbf{Resampling Update:} To ensure that the ``dirty sample'' can be sampled in the future to avoid falling into a local optimum, and to prevent it from being sampled again in the short term, we designed a replaceable sampler and reduced the sampling rate of ``dirty samples'', defined as:
\begin{equation}
    P(i_d)_{\text{new}} = \gamma \cdot P(i_d)_{\text{old}}, \\
\label{eq:resampling_weight}
\end{equation}
where $\gamma$  is the hyperparameter representing the resampling down-weighting factor, set to 0.8. As shown in Figure \ref{fig:LISA_test accuracy comparison}, on the complex visual reasoning test set LISA, accuracy improves with increasing response length, drops sharply once the model collapses.

\subsection{Adaptive Difficulty Sampling (ADS)}
\label{sec:ads_offline_curation}

GRPO's random sampling (no replacement) can curb learning key current-state knowledge. Online course learning, though useful, risks local optima and higher compute costs. To boost efficiency and evade local optima, we introduce a lossless two-stage dynamic learning schedule: 1) Offline Dataset Curation, which forms an initial training set stressing hard samples, and 2) Adaptive Difficulty Sampling (ADS), which tunes sampling focus by model performance.

\paragraph{Offline Dataset Curation}
\label{subsec:offline_curation_base_model}
We tested the base model Qwen 2.5-VL-3B on the RefCOCO/+/g training set, achieving 87.8\% accuracy, indicating that the pre-trained model already covers most of the training data. We then processed the 320k samples offline, dividing them into 44k difficult and 276k simple samples based on the base model's performance. These samples were randomly combined in a 1:2 ratio to create a new training dataset, enhancing learning effectiveness.

\paragraph{Adaptive Difficulty Sampling (ADS)}
\label{subsec:ads}
Based on the reorganized training data, we designed the Adaptive Difficulty Sampling (ADS) online learning schedule. After cleaning the ``dirty samples'' as described in Section \ref{sec:RRS}, we classify normal samples into easy, moderate, and hard categories based on the accuracy reward $R_{acc}^{i}$ from the current policy $\pi_\theta$. For easy samples, we reduce the sampling rate and allow normal gradient backpropagation; for hard samples, we reduce the sampling rate and prohibit backpropagation; for moderate samples, we increase the sampling rate and allow normal backpropagation. These adjustments are defined as:

\begin{equation}
    i = \begin{cases}
        i_{easy} & \text{if } R_{acc}^{(i)} > \theta_{H} \\
        i_{hard} & \text{if } R_{acc}^{(i)} < \theta_{L} \\
        i_{moderate} & \text{if } \theta_{L} \le R_{acc}^{(i)} \le \theta_{H},
    \end{cases}   
    \quad \quad \quad 
    P_{i} = \begin{cases}
        \alpha_{easy} \cdot P_{i} & \text{if } i = i_{easy} \\
        \alpha_{hard} \cdot P_{i} & \text{if } i = i_{hard} \\
        \alpha_{moderate} \cdot P_{i} & \text{if } i = i_{moderate}.
    \end{cases}
\label{eq:ADS}
\end{equation}
Here, $\theta_{H}$ and $\theta_{L}$ are set to 0.5 and 0.2, respectively, while $\alpha_{easy}$, $\alpha_{hard}$ and $\alpha_{moderate}$ are set to 0.1, 0.8, and 1.5, respectively. This approach eliminates the possibility of resampling easy samples, prevents overly difficult samples from interfering, and allows moderate difficulty samples to be learned multiple times until mastered, similar to how humans gradually learn moderate-level knowledge while minimizing the impact of overly challenging problems.

\subsection{Test-Time Resolution Scaling (TTRS)}
\label{sec:test_time_scaling}
RL methods require high GPU memory, making it difficult to use high-resolution images during training, leading many public datasets to provide compressed images. However, real-world testing often involves varying image resolutions, creating a gap between training and testing sets that can affect model performance. To address this while minimizing computational overhead, we propose the Test-Time Resolution Scaling (TTRS).

At inference, TTRS standardizes the input image dimensions by resizing the shorter side of the original image  $(W_{\text{orig}}, H_{\text{orig}})$ to a target length $S_{\text{target}}$ (672 pixels in this work). This resizing is done while preserving the aspect ratio to minimize geometric distortion. The size are calculated by:
\begin{equation}
W_{\text{scaled}} = \frac{W_{\text{orig}} \cdot S_{\text{target}}}{\min(W_{\text{orig}}, H_{\text{orig}})}, \quad H_{\text{scaled}} = \frac{H_{\text{orig}} \cdot S_{\text{target}}}{\min(W_{\text{orig}}, H_{\text{orig}})}.
    \label{eq:scaled_dimensions}
\end{equation}

\paragraph{Test-Time Multi-Scale Ensemble}
\label{subsec:multi_scale_ensemble}

\begin{wrapfigure}{r}{0.45\textwidth} 
    \centering
    \vspace{-4.5mm} 
    \renewcommand{\arraystretch}{1.1}
    \setlength{\tabcolsep}{5pt} 
    \captionof{table}{Example of model performance variation with input different scales in LISA.
    }
    \label{tab:scale_sensitivity_example}
    \scalebox{0.85}{ 
        \begin{tabular}{c|c}
            \hline
            \textbf{Scale (Short Side)} & \textbf{Accuracy (\%)} \\
            \hline
            w/o TTRS & 63.51 \\
            560px & 69.55 \\
            672px & 67.19 \\
            800px & 68.68 \\
            w/ TTME & 70.81 \\
            \hline
        \end{tabular}
    } 
    \vspace{-3mm} 
\end{wrapfigure}

As shown in Table~\ref{tab:scale_sensitivity_example}, the model is sensitive to input image scale in visual reasoning, with predictions varying across different scales. Analysis of these multi-scale predictions reveals that the model often deviates from the optimal solution, with the ``correct answer'' concentrated in a few key scales. Based on this insight, to further enhance model performance, we propose the Test-Time Multi-Scale Ensemble (TTME) strategy. This method first processes the input image into $\beta$ different scales, calculates $\beta$ accuracy rewards through the model, and selects the answer with the least number of intersections as the final answer. In this work, $\beta$ is set to 3. For REC, we calculate the answer with the least IoU overlap, and for VQA, we select the answer with the least inconsistency or the longest edit distance. Notably, using single-shot TTRS accelerates inference speed (e.g., 66.7\% faster on the LISA test set) by compressing the test sample resolution, while TTME has nearly lossless inference time. A detailed computational efficiency analysis will be provided in the supplementary material.

\section{Experiments}
\label{others}

\subsection{Setup}

\textbf{LVLMs}. Qwen 2.5-VL-3B serves as our base model, selected for its promising capabilities in vision-language understanding, which we aim to further enhance using reinforcement learning.

\textbf{Training datasets for REC.} To evaluate the generalization of foundational REC skills to advanced reasoning, our model trains on the RefCOCO/+/g splits \cite{mao2016generation, yu2016modeling}, which focus on visual attributes like object location and appearance rather than multi-step or abstract reasoning. \textbf{Evaluation datasets for REC.}  ID performance is measured on the validation and test splits of RefCOCO/+/g \cite{mao2016generation, yu2016modeling}. For OOD generalization, we use RefGTA \cite{tanaka2019generating} to test visual domain shift with synthetic human images, and the LISA-Grounding test split \cite{lai2024lisa} to assess reasoning transfer in tasks requiring fine-grained visual-linguistic and relational understanding.

\textbf{Training datasets for VQA.} For VQA training, we compiled a dataset of 9,600 instances by randomly sampling from multiple sub-datasets in the R1-Vision collection \cite{Shen2024R1Onevision}, including MathQA, ChartQA, DeepForm, DocVQA, InfographicsVQA, TextVQA, and OCRVQA. This sample size aligns with our 800-step training procedure. \textbf{Evaluation datasets for VQA.} VQA performance is evaluated using specialized datasets such as MMStar \cite{chen2024we}, AI2D \cite{kembhavi2016diagram}, InfoVQA VAL \cite{mathew2022infographicvqa}, TextVQA VAL \cite{singh2019towards}, DocVQA VAL \cite{mathew2021docvqa}, MATH-Vision-FULL \cite{wang2024measuring}, and MMBench \cite{liu2024mmbenchmultimodalmodelallaround}, testing the model’s capabilities across various VQA tasks.

\textbf{Baseline.} We use VLM-R1 \cite{shen2025vlm}, a framework specifically designed to enhance visual reasoning capabilities of LVLMs through reinforcement learning for comparision.

\subsection{Experimental Results}

\paragraph{Comparison to State of the Art}

\newcolumntype{T}{!{\vrule width 0.7pt}} 

\begin{table}[t]
\caption{Accuracy of state-of-the-art MLLM models on ID visual grounding benchmarks. The best performance is reported here for each method. Our method achieves the best accuracy in most cases.}
\vspace{2mm}
\renewcommand{\arraystretch}{1.2}
\label{tab:my-table_grounding_id}
\setlength{\tabcolsep}{6pt} 
\scalebox{0.62}{ 
\begin{tabular}{c T c|c|c T c|c|c T c|c T c}
\hline
\multirow{2}{*}{\textbf{Model}} & \multicolumn{3}{c T}{RefCOCO} & \multicolumn{3}{c T}{RefCOCO+} & \multicolumn{2}{c T}{RefCOCOg} & \textbf{Avg.}\\
\cline{2-10}
~ & val & testA & \multicolumn{1}{c T}{testB} & val & testA & testB & val & test &  \\
\hline

\hline
Grounding DINO-Tiny~\cite{liu2023hidden} & 89.2 & 91.9 & 86.0 & 81.1 & 87.4 & 74.7 & 85.2 & 84.9 & 85.1 \\
Grounding DINO-Largey~\cite{liu2023hidden} & 90.6 & 93.2 & 88.2 & 82.8 & 89.0 & 75.9 & 86.1 & 87.0 & 86.6\\
HieA2G~\cite{wang2025hierarchical} & 87.8 & 90.3 & 84.0 & 80.7 & 85.6 & 72.9 & 83.7 & 83.8 & 83.6\\
InternVL2-1B~\cite{team2024internvl2} & 83.6 & 88.7 & 79.8 & 76.0 & 83.6 & 67.7 & 80.2 & 79.9 & 79.9\\
InternVL2-2B~\cite{team2024internvl2} & 82.3 & 88.2 & 75.9 & 73.5 & 82.8 & 63.3 & 77.6 & 78.3 & 77.7\\
Qwen2.5VL-3B~\cite{bai2025qwen2} & 88.4 & 91.2 & 83.6 & 80.6 & 86.9 & 72.8 & 84.2 & 84.7 & 84.1\\
VLM-R1(trained 1000-step) \cite{shen2025vlm} & 90.3 & 92.5 & 85.7 & 84.3 & 89.4 & 77.1 & 86.1 & 86.8 & 86.5\\
Ours(trained 1000-step) & 91.8(+3.4) & 93.4(+2.2) & 87.7(+4.1) & 86.3(+5.7) & 90.8(+3.9) & 79.7(+6.9) & 87.8(+3.6) & 88.3(+3.6) & 88.2(+4.1)\\
\hline
\end{tabular}
} 
\vspace{-5mm}
\end{table}

\textbf{REC.} We compare our TACO method with top-performing models on the in-domain (ID) dataset RefCOCO/+/g and out-of-domain (OOD) datasets RefGTA and LISA. Table~\ref{tab:my-table_grounding_id} shows that TACO surpasses the baseline VLM-R1 by +1.7\%, outperforms Qwen 2.5-VL-3B by +4.1\%, and exceeds specialized models like Grounding DINO-Large by +1.6\%, demonstrating TACO's strong visual reasoning in ID scenarios. Notably, we trained for only 1,000 steps with 6 samples per step, using a small fraction (1.875\%) of the 320k unique region descriptions in RefCOCO. Table~\ref{tab:my-table_grounding_ood_wrap} shows TACO achieving 75.1\% accuracy on LISA (19.7\% higher than Qwen 2.5-VL-3B) and 78.7\% on RefGTA (7.9\% higher than the base model). These results highlight TACO’s strong performance in complex reasoning OOD tasks, showcasing its generalization capabilities.

\begin{wrapfigure}{r}{0.45\textwidth}
\centering
\vspace{-4mm}
\renewcommand{\arraystretch}{1.1}
\setlength{\tabcolsep}{5pt}
\captionof{table}{Performance (accuracy) comparison on OOD Benchmark.} 
\label{tab:my-table_grounding_ood_wrap}
\scalebox{0.8}{
\begin{tabular}{c T c | c}
\hline
\multirow{2}{*}{\textbf{Model}} & \multicolumn{1}{c|}{LISA} & \multicolumn{1}{c}{RefGTA} \\
\cline{2-3}
~ & val & val \\
\hline
Qwen2.5VL-3B~\cite{bai2025qwen2} & 55.4 & 70.8 \\
VLM-R1 \cite{shen2025vlm} & 61.2 & 71.6 \\
Ours & 66.5(+10.6) & 74.9(+4.1) \\
Ours(w TTME) & 75.1(+19.7) & 78.7(+7.9) \\
\hline
\end{tabular}
} 
\vspace{-4mm}
\end{wrapfigure}

The results in Table~\ref{table:rec results} clearly demonstrate that our method surpasses both SFT and VLM-R1 in continuous learning capabilities and out-of-distribution (OOD) generalization.
On in-domain datasets (RefCOCO/+/g), SFT showed minimal average accuracy improvement after 800 training steps (approximately 0.34 points), whereas our method achieved a significant average increase of about 3.30 points.
Compared to VLM-R1, which improved by an average of approximately 2.06 points, our method performed better by an average of 1.24 points on these in-domain tasks, showcasing superior learning potential and a higher performance ceiling.

On the challenging OOD LISA-Grounding dataset, our method's advantages are particularly striking: after 800 training steps, its accuracy exceeded SFT by 11.61 points and VLM-R1 by 5.34 points (see $\Delta$ rows in Table~\ref{table:rec results}).
These figures strongly attest to the effectiveness of our approach in enhancing model stability and learning potential. Furthermore, as shown in other experimental results (e.g., Table~\ref{tab:my-table_grounding_ood_wrap}), the TTRS strategy further boosts performance on OOD tasks, highlighting the efficacy of TTME in improving generalization.

\definecolor{darkgreen}{RGB}{0, 150, 0}
\begin{table*}[htbp]
\scriptsize
  \begin{center}
  \setlength{\tabcolsep}{11pt} 
  \vspace{-1mm}
  \caption{Performance (accuracy) comparison of SFT and RL methods on ID and OOD benchmarks. Scores for RefCOCO/+/g represent average accuracies across sub-datasets (see Appendix for details). All models are based on Qwen 2.5-VL-3B, with SFT and RL using RefCOCO/+/g training splits. Scores at ``Step 0'' correspond to the Qwen 2.5-VL-3B model. $\Delta_{RL-SFT}$ represents the RL model's gain over SFT, and $\Delta_{RL-VLM-R1}$ shows our model's advantage over VLM-R1.}
  \label{table:rec results}
  \renewcommand{\arraystretch}{1.1}
    \begin{tabular}{c c c c c c c c c}
    \toprule
        \textbf{Training} & &  \textbf{Evaluation} &  \multicolumn{6}{c}{\textbf{Training Steps}}\\
      \textbf{Method} &  & \textbf{Dataset} & & 0 & 200 & 400 & 600 &  800 \\
    \midrule
        SFT & \vline & \multirow{3}{*}{Refcoco} & \vline & 87.79 & 88.08 & 88.16 & 88.21 & 88.27 \\ 
        VLM-R1 & \vline &  &  \vline & 87.79 & 88.80 & 89.12 & 89.30 & 89.42 \\ 
        Ours & \vline &  &  \vline & 87.79 & 89.55 & 89.96 & 90.36 & 90.42 \\ 
    \noalign{\vskip 0.25ex}
    \hdashline
    \noalign{\vskip 0.25ex}
        SFT & \vline & \multirow{3}{*}{Refcoco+} & \vline & 80.63 & 81.60 & 81.31 & 81.19 & 81.28 \\ 
        VLM-R1 & \vline &  &  \vline & 80.63 & 82.39 & 82.64 & 83.26 & 83.50 \\ 
        Ours & \vline &  &  \vline & 80.63 & 83.45 & 84.32 & 84.67 & 85.00 \\ 
    \noalign{\vskip 0.25ex}
    \hdashline
    \noalign{\vskip 0.25ex}
        SFT & \vline & \multirow{3}{*}{Refcocog}  &  \vline& 84.79 & 85.02 & 84.80 & 84.59 & 84.68 \\ 
        VLM-R1 & \vline &  &  \vline& 84.79 & 85.36 & 85.86 & 86.38 & 86.46 \\ 
        Ours & \vline &  &  \vline & 84.79 & 86.57 & 87.31 & 87.40 & 87.70 \\ 
    \midrule
        SFT & \vline & \multirow{5}{*}{LISA-Grounding}  &  \vline& 55.37 & 56.15 & 54.95 & 54.16 & 54.83 \\ 
        VLM-R1 & \vline & \multirow{3}{*}{}    &  \vline& 55.37 & 61.76 & 62.00 & 60.68 & 61.10 \\ 
        Ours & \vline & \multirow{3}{*}{}      &  \vline & 55.37 & 62.97 & 64.49 & 65.26 & 66.44 \\  
        \cdashline{5-9}
        \noalign{\vskip 0.25ex}
        $\Delta_{Ours-SFT}$ & \vline &   &  \vline & 0 & \textcolor{darkgreen}{+6.82} & \textcolor{darkgreen}{+9.54} & \textcolor{darkgreen}{+11.10} & \textcolor{darkgreen}{+11.61} \\
        \cdashline{5-9}
        \noalign{\vskip 0.25ex}
        $\Delta_{Ours-VLM-R1}$ & \vline &   &  \vline & 0 & \textcolor{darkgreen}{+1.21} & \textcolor{darkgreen}{+2.49} & \textcolor{darkgreen}{+4.58} & \textcolor{darkgreen}{+5.34} \\
    \bottomrule
    \end{tabular}
  \end{center}

\end{table*}

\begin{table}[h]
\centering
\setlength{\tabcolsep}{10pt}
\caption{Comparison results of our method and Qwen2.5VL-3B on various multimodal benchmarks.}
\vspace{2mm}
\renewcommand{\arraystretch}{1.2} 
\label{tab:main_benchmarks_selected} 
\scalebox{0.7}{ 
\begin{tabular}{l T c T cccc T c | c} 
\hline
\multicolumn{1}{lT}{\multirow{3}{*}{\textbf{Model}}} &
\multicolumn{1}{cT}{\textbf{Math}} & 
\multicolumn{6}{c}{\textbf{General Visual Question Answering}} \\
\cline{3-8} 
~ & 
\multicolumn{1}{cT}{\textbf{Vision}} & 
\multicolumn{4}{cT}{MMBench} & 
\multicolumn{1}{cT}{MMStar} & 
\multicolumn{1}{c}{AI2D} \\    
\cline{2-2} \cline{3-8} 
~ & 
\multicolumn{1}{cT}{\textbf{(Full)}} & 
EN(dev) & CN(dev) & EN-V11(dev) & \multicolumn{1}{cT}{CN-V11(dev)} & 
\multicolumn{1}{cT}{(test)} & 
\multicolumn{1}{c}{(test)} \\   
\hline 
Qwen2.5VL-3B & 20.1 & 78.0 & 77.2 & 75.8 & 75.6 & \multicolumn{1}{cT}{53.0} & 77.4 \\
\textbf{Ours} & 24.1 (+4.0) & 81.1 (+3.1) & 79.0 (+1.9) & 79.3 (+3.5) & 78.0 (+2.4) & \multicolumn{1}{cT}{59.9 (+6.9)} & 80.5 (+3.1) \\
\hline
\end{tabular}
} 
\vspace{-1mm}
\end{table}

\begin{wrapfigure}{r}{0.5\textwidth} 
\centering
\captionof{table}{Performance on OCR-related Understanding Tasks (InfoVQA, TextVQA, DocVQA). The best accuracy is reported here for each method.} 
\renewcommand{\arraystretch}{1.2}
\label{tab:combined_ocr_tasks_all_models_wrap_citations} 
\scalebox{0.7}{ 
\begin{tabular}{l T c T c c} 
\hline
\multicolumn{1}{lT}{\multirow{2}{*}{\textbf{Model}}} & 
\multicolumn{1}{cT}{\textbf{InfoVQA}} & 
\multicolumn{1}{cT}{\textbf{TextVQA}} & 
\multicolumn{1}{c}{\textbf{DocVQA}} \\
\cline{2-4} 
~ & 
\multicolumn{1}{cT}{(VAL)} & 
\multicolumn{1}{cT}{(VAL)} & 
\multicolumn{1}{c}{(VAL)} \\
\hline 
LLaVA-OV 0.5B~\cite{li2024llava} & 41.8 & \multicolumn{1}{cT}{-} & 70.0 \\
InternVL2-1B~\cite{team2024internvl2} & 50.9 & \multicolumn{1}{cT}{70.5} & 81.7 \\
MM 1.5-1B~\cite{zhang2024mm1} & 50.5 & \multicolumn{1}{cT}{72.5} & 81.0 \\
MolmoE-1B~\cite{deitke2024molmo} & 53.9 & \multicolumn{1}{cT}{78.8} & 77.7 \\
MiniCPM-V 2.0~\cite{yao2024minicpm} & - & \multicolumn{1}{cT}{74.1} & 71.9 \\
InternVL2-2B~\cite{team2024internvl2} & 58.9 & \multicolumn{1}{cT}{73.4} & 86.9 \\
Qwen2-VL-2B~\cite{wang2024qwen2} & 65.5 & \multicolumn{1}{cT}{\textbf{79.7}} & 90.1 \\
MM 1.5-3B~\cite{zhang2024mm1} & 58.5 & \multicolumn{1}{cT}{76.5} & 87.7 \\
Qwen2.5VL-3B & 75.1 & \multicolumn{1}{cT}{78.7} & \textbf{93.0} \\
\textbf{Ours} & \textbf{77.6(+2.5)} & \multicolumn{1}{cT}{79.0(+0.3)} & 92.6(-0.4) \\
\hline
\end{tabular}
} 
\end{wrapfigure}

\textbf{VQA.} Our model exhibits significant performance gains and strong generalization across diverse benchmarks. On general multimodal tasks (Table~\ref{tab:main_benchmarks_selected}), it outperforms Qwen2.5VL-3B, showing key improvements such as +6.9\% on MMStar (test) and +4.0\% on Math Vision (Full), along with consistent gains on MMBench and AI2D. This adaptability is further demonstrated on OCR-specific benchmarks (Table~\ref{tab:combined_ocr_tasks_all_models_wrap_citations}), where our model leads with 77.6\% on InfoVQA and 79.0\% on TextVQA, while maintaining strong performance on DocVQA with 92.6\%. These results collectively underscore our model’s ability to efficiently develop broad and robust capabilities through its versatility. Our single model for mixed VQA tasks shows improved generalization and stability over specialized ones.

\paragraph{Ablation Studies}
As illustrated in Table ~\ref{table:ablation}, we observe that for the LISA dataset, each algorithm component contributes significantly to the overall performance, with TTRS making the largest contribution. This is due to the substantial resolution distribution difference between the LISA test set and the training data. In contrast, for the RefCOCO/+/g datasets, all algorithm components, except for TTRS, contribute significantly to the overall performance. This is because RefCOCO/+/g, as an in-domain test set, does not exhibit a noticeable resolution distribution difference.

\begin{table}[h!]
\vspace{-3mm}
\centering
\caption{Ablation results of our method on different datasets.}
\vspace{2mm}
\label{table:ablation}
\small
\setlength{\tabcolsep}{5pt}
\begin{tabular}{|l|c|c|c|c|c|c|c|}
\hline 
 \textbf{TAC} & \textbf{RRS} & \textbf{ADS} & \textbf{TTRS} & \textbf{RefCOCO} & \textbf{RefCOCO+} & \textbf{RefCOCOg}  & \textbf{LISA}\\
\hline 
  & & & & 89.5 & 83.6 & 86.5 & 61.2\\
\checkmark   &  & & & 90.1 & 84.6  & 87.1  & 63.2\\
\checkmark & \checkmark &  &   & 90.5 & 85.1  & 87.4  & 65.1\\
\checkmark & \checkmark &  \checkmark&   & 90.8 & 85.6  & 87.6    & 66.5\\
\checkmark & \checkmark &  \checkmark&  \checkmark   & 90.8 & 85.5  & 87.6  & 75.1\\
\hline 
\end{tabular}
\vspace{-3mm}
\end{table}

\section{Limitations}
Although our TACO method has achieved significant performance improvements, it still struggles to correctly infer some unknown samples due to the lack of world knowledge. Additionally, it faces challenges in making accurate predictions for visual problems such as occlusion and blur. We provide examples of these cases in the supplementary materials.

\section{Conclusion}
In this work, we propose TACO, a novel reinforcement learning algorithm for LVLMs that addresses key challenges in visual reasoning, including inconsistencies in reasoning, model instability, and low data efficiency. By incorporating Think-Answer Consistency, Rollback Resample Strategy, and an adaptive learning schedule, TACO enhances model stability and learning efficiency. Additionally, the Test-Time-Resolution-Scaling scheme mitigates performance degradation caused by varying resolutions. Extensive experiments show that TACO achieves significant performance improvements on both in-distribution and out-of-distribution benchmarks for REC and VQA tasks, demonstrating its strong generalization and versatility.

\bibliographystyle{unsrt}  
\small
\bibliography{ref}
\normalsize

\end{document}